\documentclass[conference]{IEEEtran}
\IEEEoverridecommandlockouts
\usepackage{cite}
\usepackage{amsmath,amssymb,amsfonts}
\usepackage{algorithmic}
\usepackage{graphicx}
\usepackage{textcomp}
\usepackage{xcolor}
\usepackage{import}
\usepackage{hyperref}
\usepackage{mathtools}
\usepackage{commath}
\usepackage{fixmath}
\usepackage{siunitx}
\usepackage{ntheorem}
\usepackage{subcaption}
\def\BibTeX{{\rm B\kern-.05em{\sc i\kern-.025em b}\kern-.08em
    T\kern-.1667em\lower.7ex\hbox{E}\kern-.125emX}}
\begin{document}

\def \hrsup_size{0.48}
\def \framesu_size{0.48}
\newtheorem*{proposition}{Proposition}

\title{Trajectory Advancement for Robot Stand-up \\ with Human Assistance
\thanks{This work is supported by \href{http://itn-pace.eu/}{PACE} project, Marie Skłodowska-Curie grant agreement No. 642961 and \href{https://andy-project.eu/}{An.Dy} project which has received funding from the European Union\textquotesingle s Horizon 2020 Research and Innovation Programme under grant agreement No. 731540.}
}

\author{\IEEEauthorblockN{Yeshasvi Tirupachuri, Gabriele Nava, Lorenzo Rapetti, Claudia Latella, Daniele Pucci}
\IEEEauthorblockA{\textit{Dynamic Interaction Control}, \textit{Italian Institute of Technology}, Genova, Italy \\
{\tt\small name.surname@iit.it}}
}

\maketitle

\begin{abstract}


Physical interactions are inevitable part of human-robot collaboration tasks and rather than exhibiting simple reactive behaviors to human interactions, collaborative robots need to be endowed with intuitive behaviors. This paper proposes a trajectory advancement approach that facilitates advancement along a reference trajectory by leveraging assistance from helpful interaction wrench present during human-robot collaboration. We validate our approach through experiments in simulation with iCub.

\end{abstract}


\section{Introduction}
\label{sec:introduction}

Technological progress is headed in the direction of co-existence between humans and robots. The research field of Human-Robot Collaboration (HRC) is vital to realize many potential applications such as collaborative manufacturing and elderly assistance. Typical HRC scenarios involve a human and a robotic agent engaged in physical interactions with a common goal of accomplishing a task. During such scenarios, rather than exhibiting simple reactive robot behavior it is desirable to display intuitive behavior leveraging human assistance to achieve the robot's task quicker. This paper presents an intuitive approach of advancement along a reference trajectory by exploiting assistance provided during HRC.

Physical interactions during HRC are often intentional and can provide informative insights that can augment the task completion \cite{bajcsy2017learning}.
Consider an example case of a robot moving its center of mass (CoM) along a given Cartesian reference trajectory to perform a complicated task of sit-to-stand transition. An intuitive interaction of a human with the intention to speed up the robot motion is to apply forces in the robot's desired direction. Under such circumstances, traditionally, the robot can either render a compliant behavior through impedance/admittance control or be robust to any external interactions even if it is helpful for the task at hand. Instead, a more intuitive behavior is to advance further along the reference trajectory and stand-up quicker. This motivates us to propose a trajectory advancement approach through which the robot can advance along the reference trajectory leveraging assistance from physical interactions.

\section{BACKGROUND}
\label{sec:background}

The equations of motion of a \textit{floating base} robotic system are described by,

\begin{equation}
	M(q) \dot{\nu} + C(q,\nu) \nu + G(q) = B {\tau} + J_c^T f^*
	\label{eq:equations-of-motion}
\end{equation}

where, $M \in \mathbb{R}^{n+6 \times n+6}$ is the mass matrix, $C \in \mathbb{R}^{n+6 \times n+6}$ is the Coriolis matrix, $G \in \mathbb{R}^{n+6}$ is the gravity term, $B = (0_{n \times 6},1_n)^T$ is a selector matrix, ${\tau}  \in \mathbb{R}^{n}$ is a vector representing the robot's joint torques, $f^* \in \mathbb{R}^{6n_c}$ represents the external wrenches acting on $n_c$ contact links of the robot, and $J_c \in \mathbb{R}^{n+6 \times 6n_c}$ is the contact jacobian. Consider the problem of Cartesian trajectory tracking by a link of the robot where $x_d(t), \dot{x}_d(t), \ddot{x}_d(t) \in \mathbb{R}^6$ denote the desired position, velocity and acceleration in Cartesian space, parametrized in time $t$. Now, $\dot{\widetilde{x}} = \dot{x}(t) - \dot{x}_d(t)$ is the velocity tracking error to be minimized. The control objective for the tracking task is defined as,

\begin{equation}
	\label{eq:cartesian-control-objective}
	\ddot{x} = \ddot{x}^* := \ddot{x}_d - K_D \ \dot{\widetilde{x}} - K_P \int_0^t \dot{\widetilde{x}} du, \quad K_D, K_P > 0
\end{equation}
where $K_P, K_D \in \mathbb{R}^{6 \times 6}$ are positive symmetric feedback matrices. According to the classical feedback linearisation approach \cite{khalil2004modeling}, the robot control torques necessary for trajectory tracking with the desired dynamics directed by Eq.~\eqref{eq:cartesian-control-objective} are given by,

\begin{equation}
    {\tau} = \mathbold{\Delta}^{\dagger} [\ddot{x}^*  - \mathbold{\Omega} f^* + \mathbold{\Lambda}]
	\label{eq:normal-control-torques-compact}
\end{equation}

where $ \mathbold{\Delta} = J M^{-1} B \in \mathbb{R}^{6 \times \mathrm{n}} $; $ \mathbold{\Omega} = J M^{-1} J_c^T \in \mathbb{R}^{6 \times 6\mathrm{n}_c}$; $ \mathbold{\Lambda} = J \ M^{-1} h - \dot{J} \nu\in \mathbb{R}^{6}$. The above control torques completely cancel out any external wrench applied during physical interactions with the robot. Although this approach is quite robust to external perturbations, it is also limited in facilitating HRC scenarios that require active collaboration between a human partner and a robot \cite{8093992}.

\section{METHOD}
\label{sec:method}

 We design a parametric curve parametrized with a \textit{free parameter} $\psi \in [0, \infty)$. The resulting parametric curve $x_d(\psi)$ is the desired geometric path to be followed spatially by a link of the robot. The term $\mathbold{\Omega} f^*$ in the control torques Eq.~\eqref{eq:normal-control-torques-compact} represents the Cartesian resultant acceleration that results under the influence of external interaction wrench $f^*$. Let us define the \emph{helpful} interaction by decomposing the external wrenches into \textit{parallel} and \textit{perpendicular} components along the desired velocity as,

\begin{subequations}
    \begin{equation}
        \mathbold{\Omega}f^* = \alpha \ \dot{x}_d^{\parallel} + \beta \ \dot{x}_d^{\perp} \notag
    \end{equation}
    \begin{equation}
        \dot{x}_d^{\parallel} = \frac{\dot{x}_d}{\norm{\dot{x}_d}}, \quad \alpha = \frac{\dot{x}_d^T \mathbold{\Omega} f^*}{\norm{\dot{x}_d}} \notag
    \end{equation}
\end{subequations}

where $\dot{x}_d^{\parallel} \in \mathbb{R}^{6}$ is the unit vector along the direction of the desired velocity, $\alpha \in \mathbb{R}$ is the resultant acceleration component projected along parallel direction of desired velocity. 


\begin{proposition}
\label{proposiiton-update-law}
The time evolution of the free parameter $\psi$ for trajectory advancement leveraging assistance is given by the following update rule,
\begin{equation}
    \dot{\psi} = min \left\{ \dot{\psi}_{upper}, max \left\{ 1, \frac{\dot{x}(t)^T \ \partial_{\psi} x_d(\psi)}{\norm{\partial_{\psi} x_d(\psi)}^2}  \right\}\right\}
    \label{eq:update-rule}
\end{equation}

\end{proposition}

A complete proof of the proposition and the choice of $\alpha$ is available in \cite{1907.13445}.
\section{Experiments \& Results}
\label{sec:experiments-results}

\begin{figure}[t]
    \centering
    \begin{subfigure}{0.125\textwidth}
        \centering
        \includegraphics[scale=0.10]{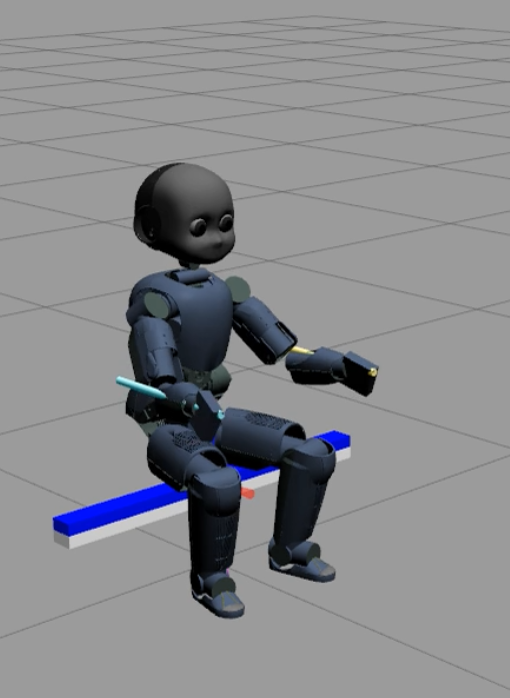}
        \caption{}
        \label{fig:state-1.png}
    \end{subfigure}%
    \begin{subfigure}{0.125\textwidth}
        \centering
        \includegraphics[scale=0.10]{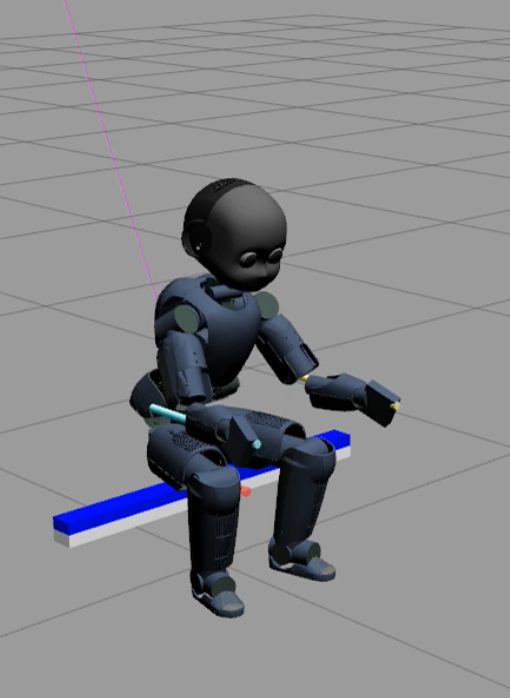}
        \caption{}
        \label{fig:state-2.png}
    \end{subfigure}%
    \begin{subfigure}{0.125\textwidth}
        \centering
        \includegraphics[ scale=0.10]{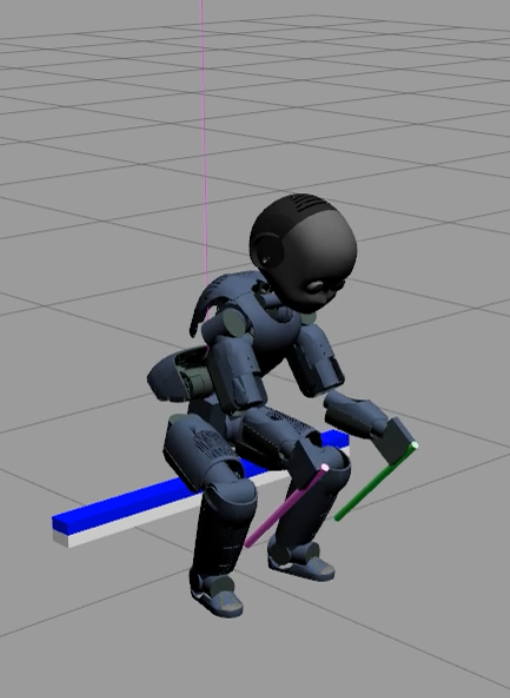}
        \caption{}
        \label{fig:state-3.png}
    \end{subfigure}%
    \begin{subfigure}{0.125\textwidth}
        \centering
        \includegraphics[scale=0.10]{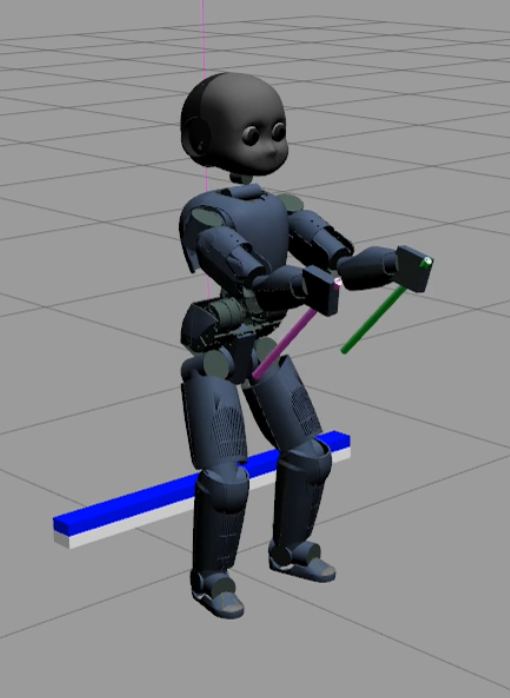}
        \caption{}
        \label{fig:state-4.png}
    \end{subfigure}
    \caption{iCub at different states during sit-to-stand transition}
    \label{fig:standup-states}
\end{figure}

The task of the robot is to do a sit-to-stand transition by moving its center of mass along a reference trajectory with momentum control as the primary objective \cite{8093992} \cite{nava2016stability}. The robotic platform used in our simulation experiments is the iCub humanoid robot \cite{Nataleeaaq1026} \cite{metta2010icub}. The external wrench that mimic the physical interactions from a human are realized through a plugin \cite{hoffman2014yarp}. The Simulink controller is designed with four states for the robot as highlighted in  Fig. \ref{fig:standup-states}. 

During the state 1, the robot balances on a chair and enters to state 2 when an interaction wrench of a set threshold is detected at the hands indicating the start of pull-up assistance from an external agent Fig. \ref{fig:state-1.png}. During state 2, the robot moves its center of mass forward and enters state 3 when the external wrench experienced at the feet of the robot is above a set threshold Fig. \ref{fig:state-2.png}. During state 3, the robot moves its center of mass both in forward and upward directions and enters state 4 when the external wrench experienced at the feet of the robot are above another set threshold Fig. \ref{fig:state-3.png}. Finally, during state 4 the robot moves its center of mass further upward to stand fully erect on both the feet Fig. \ref{fig:state-4.png}. Also, the external wrenches applied at the hands of the robot are shown.

\begin{figure}[t]
    \centering
    \begin{subfigure}{0.245\textwidth}
        \centering
        \includegraphics[scale=0.0625]{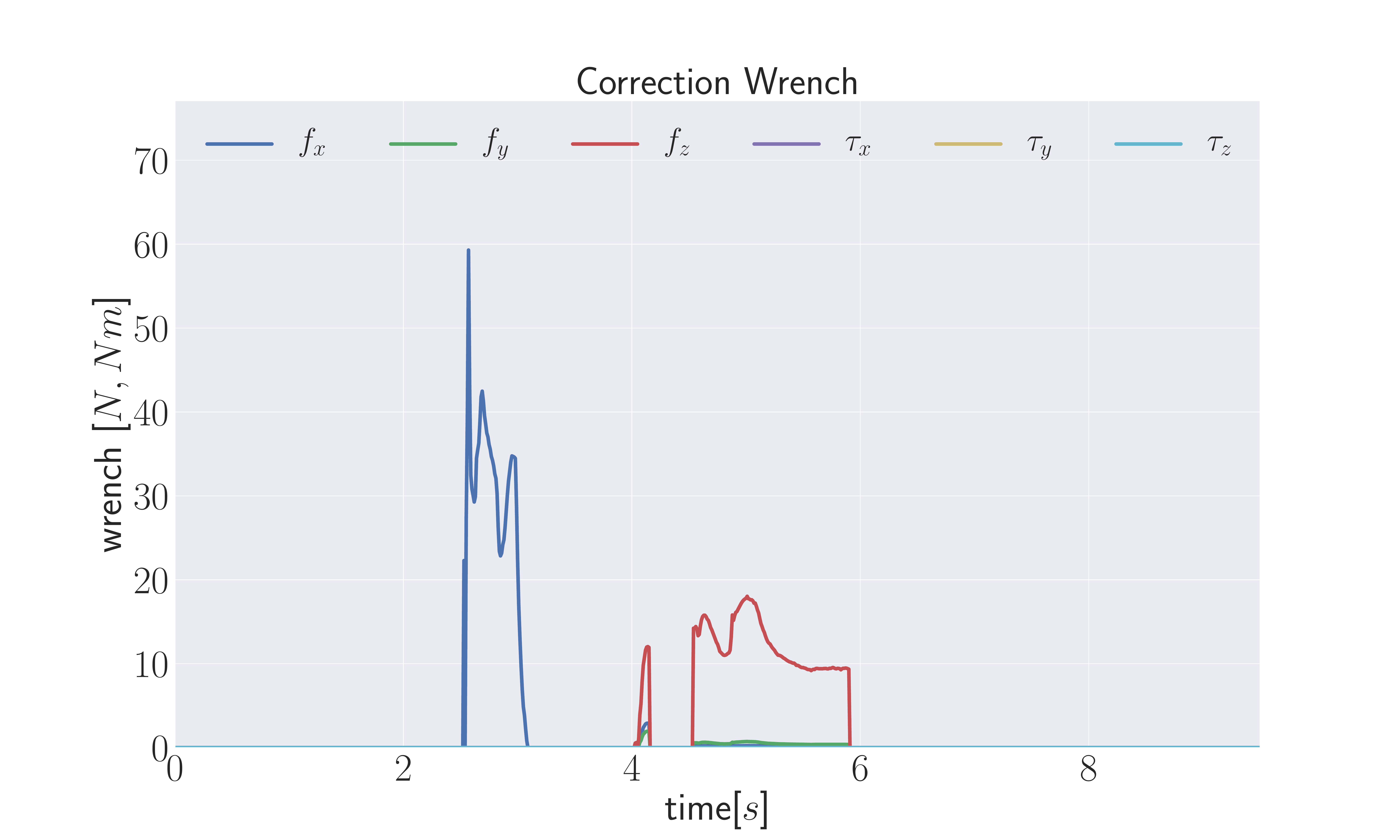}
        \caption{\hspace*{-7.5mm}}
        \label{fig:simulation-correction-wrench}
    \end{subfigure}%
    \begin{subfigure}{0.245\textwidth}
        \centering
        \includegraphics[scale=0.0625]{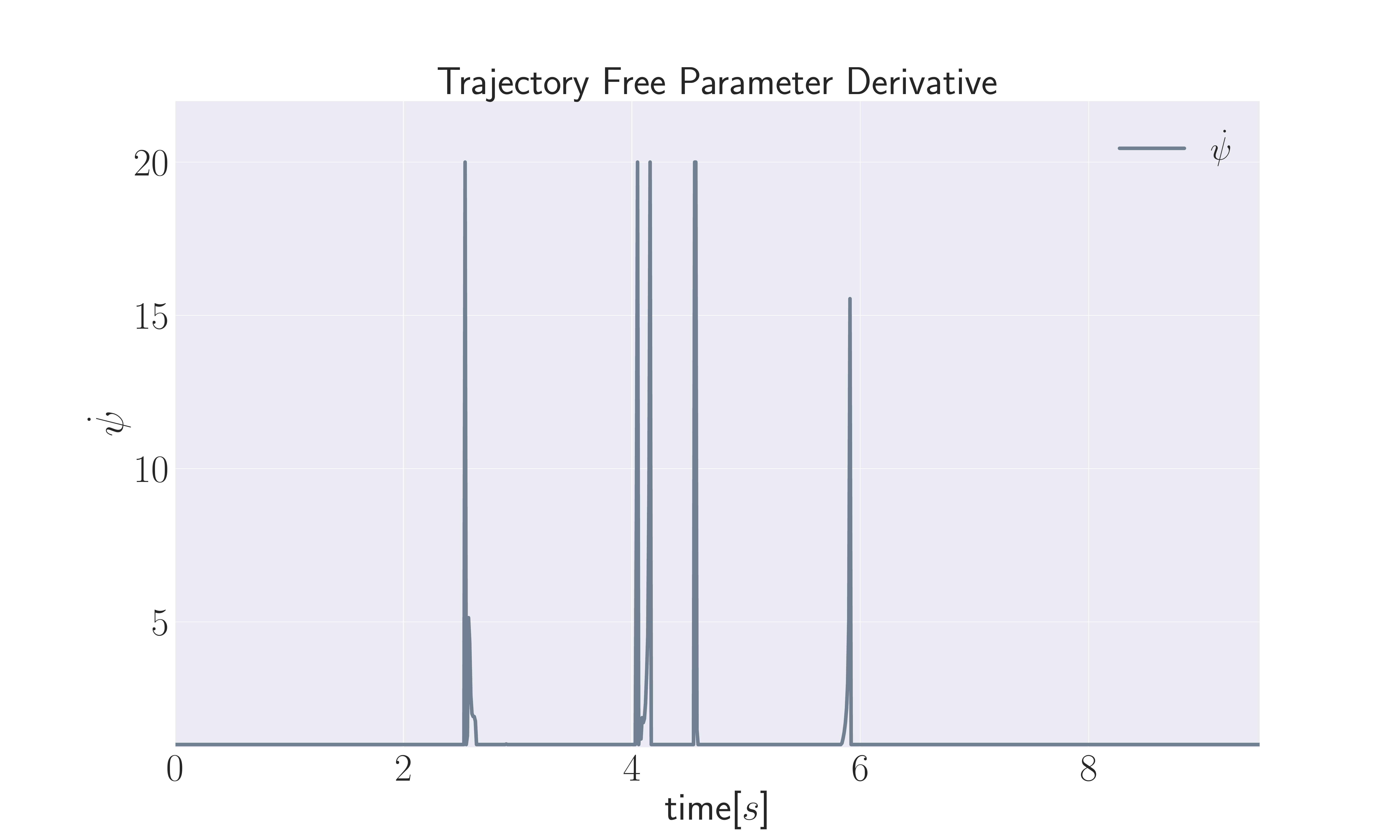}
        \caption{\hspace*{-7.5mm}}
        \label{fig:simulation-sdotvalue}
    \end{subfigure}
    \begin{subfigure}{0.245\textwidth}
        \centering
        \includegraphics[scale=0.0625]{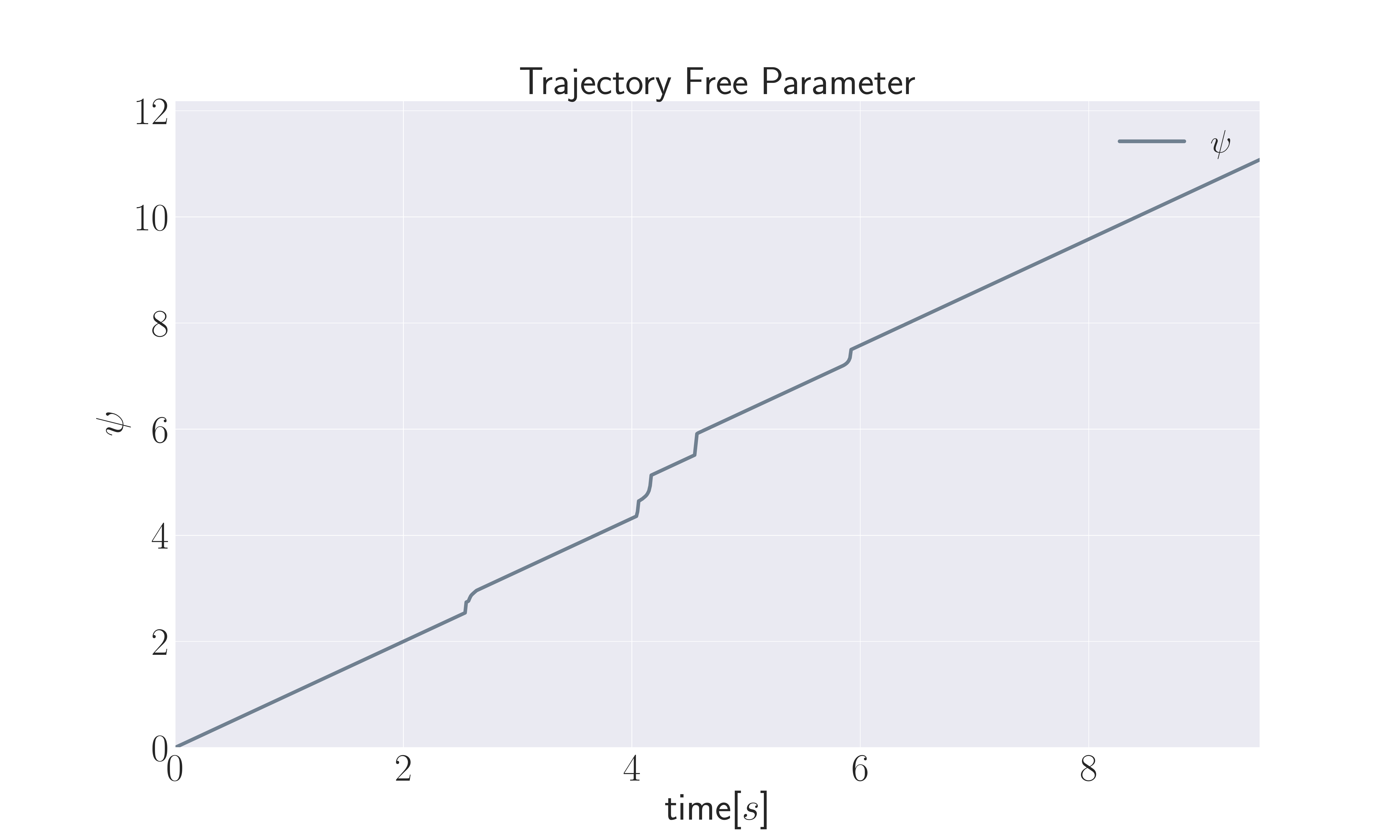}
        \caption{\hspace*{-7.5mm}}
        \label{fig:simulation-svalue}
    \end{subfigure}%
    \begin{subfigure}{0.245\textwidth}
        \centering
        \includegraphics[scale=0.0625]{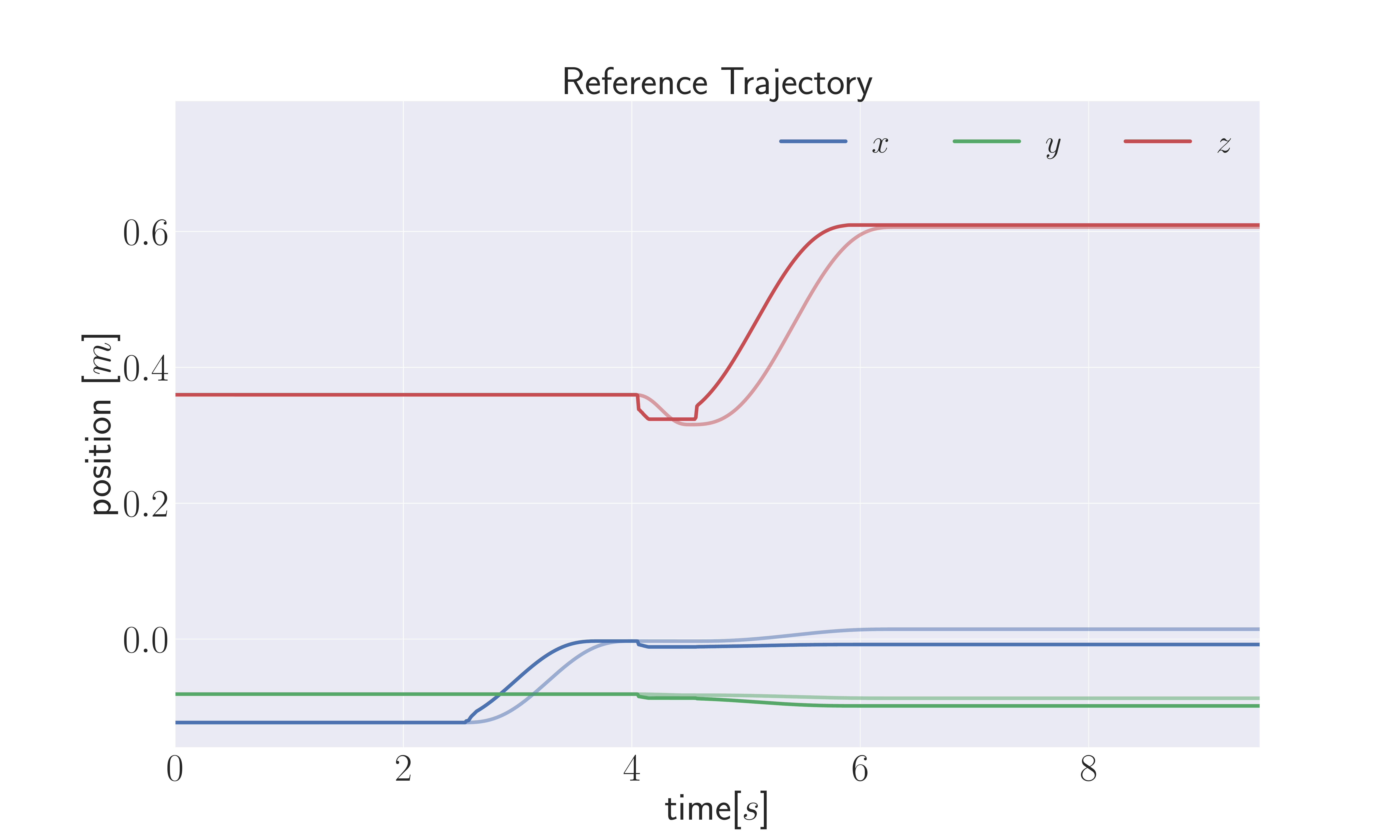}
        \caption{\hspace*{-7.5mm}}
        \label{fig:simulation-reference-trajectory}
    \end{subfigure}
    \caption{Gazebo simulation trajectory advancement during sit-to-stand transition}
    \label{fig:simulation-trajectory-advancement}
\end{figure}

The results of trajectory advancement in simulation are shown in Fig.~\ref{fig:simulation-trajectory-advancement}. The external wrench that is helpful to achieve the task is shown in Fig.~\ref{fig:simulation-correction-wrench}. The reference trajectory is similar to a time parametrized trajectory i.e., $\psi = t$ until any helpful wrench is applied at the hands of the robot. Under the influence of helpful wrenches, the derivative of the trajectory free parameter $\dot{\psi}$ changes as shown in Fig.~\ref{fig:simulation-sdotvalue} and the corresponding trajectory advancement is reflected as an increase in $\psi$ as seen in Fig.~\ref{fig:simulation-svalue}. Accordingly, the reference is advanced further along the  CoM reference trajectory as shown in Fig.~\ref{fig:simulation-reference-trajectory}. The original CoM reference trajectory without trajectory advancement is shown in the same figure with reduced transparency. 

\bibliographystyle{IEEEtran}
\bibliography{references}

\begin{thebibliography}{1}
\providecommand{\url}[1]{#1}
\csname url@samestyle\endcsname
\providecommand{\newblock}{\relax}
\providecommand{\bibinfo}[2]{#2}
\providecommand{\BIBentrySTDinterwordspacing}{\spaceskip=0pt\relax}
\providecommand{\BIBentryALTinterwordstretchfactor}{4}
\providecommand{\BIBentryALTinterwordspacing}{\spaceskip=\fontdimen2\font plus
\BIBentryALTinterwordstretchfactor\fontdimen3\font minus
  \fontdimen4\font\relax}
\providecommand{\BIBforeignlanguage}[2]{{%
\expandafter\ifx\csname l@#1\endcsname\relax
\typeout{** WARNING: IEEEtran.bst: No hyphenation pattern has been}%
\typeout{** loaded for the language `#1'. Using the pattern for}%
\typeout{** the default language instead.}%
\else
\language=\csname l@#1\endcsname
\fi
#2}}
\providecommand{\BIBdecl}{\relax}
\BIBdecl

\bibitem{bajcsy2017learning}
A.~Bajcsy, D.~P. Losey, M.~K. O’Malley, and A.~D. Dragan, ``Learning robot
  objectives from physical human interaction,'' \emph{Proceedings of Machine
  Learning Research}, vol.~78, pp. 217--226, 2017.

\bibitem{khalil2004modeling}
W.~Khalil and E.~Dombre, \emph{Modeling, identification and control of
  robots}.\hskip 1em plus 0.5em minus 0.4em\relax Butterworth-Heinemann, 2004.

\bibitem{8093992}
F.~{Romano}, G.~{Nava}, M.~{Azad}, J.~{Čamernik}, S.~{Dafarra}, O.~{Dermy},
  C.~{Latella}, M.~{Lazzaroni}, R.~{Lober}, M.~{Lorenzini}, D.~{Pucci},
  O.~{Sigaud}, S.~{Traversaro}, J.~{Babič}, S.~{Ivaldi}, M.~{Mistry},
  V.~{Padois}, and F.~{Nori}, ``The codyco project achievements and beyond:
  Toward human aware whole-body controllers for physical human robot
  interaction,'' \emph{IEEE Robotics and Automation Letters}, vol.~3, no.~1,
  pp. 516--523, Jan 2018.

\bibitem{1907.13445}
Y.~Tirupachuri, G.~Nava, L.~Rapetti, C.~Latella, and D.~Pucci, ``Trajectory
  advancement during human-robot collaboration,'' IEEE RO-MAN, in press, 2019.

\bibitem{nava2016stability}
G.~Nava, F.~Romano, F.~Nori, and D.~Pucci, ``Stability analysis and design of
  momentum-based controllers for humanoid robots,'' in \emph{Intelligent Robots
  and Systems (IROS), 2016 IEEE/RSJ International Conference on}.\hskip 1em
  plus 0.5em minus 0.4em\relax IEEE, 2016, pp. 680--687.

\bibitem{Nataleeaaq1026}
\BIBentryALTinterwordspacing
L.~Natale, C.~Bartolozzi, D.~Pucci, A.~Wykowska, and G.~Metta, ``icub: The
  not-yet-finished story of building a robot child,'' \emph{Science Robotics},
  vol.~2, no.~13, 2017. [Online]. Available:
  \url{http://robotics.sciencemag.org/content/2/13/eaaq1026}
\BIBentrySTDinterwordspacing

\bibitem{metta2010icub}
G.~Metta, L.~Natale, F.~Nori, G.~Sandini, D.~Vernon, L.~Fadiga, C.~Von~Hofsten,
  K.~Rosander, M.~Lopes, J.~Santos-Victor \emph{et~al.}, ``The icub humanoid
  robot: An open-systems platform for research in cognitive development,''
  \emph{Neural Networks}, vol.~23, no. 8-9, pp. 1125--1134, 2010.

\bibitem{hoffman2014yarp}
E.~M. Hoffman, S.~Traversaro, A.~Rocchi, M.~Ferrati, A.~Settimi, F.~Romano,
  L.~Natale, A.~Bicchi, F.~Nori, and N.~G. Tsagarakis, ``Yarp based plugins for
  gazebo simulator,'' in \emph{International Workshop on Modelling and
  Simulation for Autonomous Systems}.\hskip 1em plus 0.5em minus 0.4em\relax
  Springer, 2014, pp. 333--346.

\end{thebibliography}

\end{document}